\documentclass{article}

\usepackage{microtype}
\usepackage{graphicx}
\usepackage{subcaption}
\usepackage{booktabs} 

\usepackage{bm}
\usepackage{ltablex}
\keepXColumns
\usepackage{multirow}
\usepackage{ragged2e}
\usepackage{pdflscape}
\usepackage{soul}

\usepackage{hyperref}

\usepackage[accepted]{icml2026}

\usepackage{amsmath}
\usepackage{amssymb}
\usepackage{mathtools}
\usepackage{amsthm}

\usepackage[capitalize,noabbrev]{cleveref}
\crefformat{section}{#2Section~#1#3}
\crefformat{figure}{#2Figure~#1#3}
\crefformat{table}{#2Table~#1#3}

\theoremstyle{plain}

\theoremstyle{definition}

\theoremstyle{remark}

\usepackage[textsize=tiny]{todonotes}

\icmltitlerunning{What Does a Chemical Language Model Know About Molecules?}

\begin{document}

\twocolumn[
  \icmltitle{What Does a Chemical Language Model Know About Molecules?}



  \icmlsetsymbol{equal}{*}

  \begin{icmlauthorlist}
    \icmlauthor{Christian Kenneth}{001}
    \icmlauthor{Etowah Adams}{002}
    \icmlauthor{Liam Bai}{003}
    \icmlauthor{Gerard JP van Westen}{004}
  \end{icmlauthorlist}

  \icmlaffiliation{001}{Independent}
  \icmlaffiliation{002}{Department of Systems Biology, Columbia University, New York}
  \icmlaffiliation{003}{Generate:Biomedicines, Massachusetts}
  \icmlaffiliation{004}{Computational Drug Discovery (CDD), Division of Medicinal Chemistry, Leiden University, The Netherlands}

  \icmlcorrespondingauthor{Christian Kenneth}{kennethasikinnn@gmail.com}

  \icmlkeywords{Machine Learning, ICML}

  \vskip 0.3in
]



\printAffiliationsAndNotice{}  

\begin{abstract}
  Chemical language models (cLMs) are widely assumed to learn
  surface-level syntactic patterns rather than learning meaningful
  molecular semantics. Here, we apply sparse autoencoders (SAEs)
  to MolFormer, an encoder-only cLM, to mechanistically examine how
  molecular representations are built across layers. We discover
  that early layers rely on position-tracking latents to parse
  molecular grammar, while later layers encode atom-in-substructure
  and pharmacologically relevant features. Additionally, we show
  that non-canonical SMILES produce more disruptive representation
  shifts than invalid SMILES, driven by position-latent disruption
  propagating across layers. To support further exploration, we
  develop InterMol, an interactive visualizer for SAE activations
  on molecular strings and structures.
\end{abstract}

\section{Introduction}
  Chemical language models (cLMs, or molecular language models) have
  shown strong performance in molecular property prediction
  \cite{Ross2022-jm} and are widely used in generative drug design
  \cite{Grisoni2023-wz}. cLMs learn 'chemical language' from datasets
  containing up to billions of molecules. Some studies have further
  used compound-class-specific datasets, such as natural products
  \cite{sakano2024} and polymers \cite{Kuenneth2023-en}, to allow cLMs
  to develop specialized representations. Despite these successes,
  some in the community remain skeptical about whether cLMs learn
  meaningful chemical features or merely exploit surface-level string
  patterns, given that molecules can be represented as strings in many
  equivalent ways \cite{Bajorath2024-xq,kikuchi2026}.
  
  Addressing these concerns requires understanding how cLMs internally
  represent molecules. Recent work in AI interpretability shows that
  sparse autoencoders (SAEs)---a method for mechanistic
  interpretability---can uncover human-understandable features from
  large language models (e.g., GPT-4 \cite{gao2024} and Claude Sonnet 3
  \cite{templeton2024}) as well as domain-specific language models
  (e.g., ESM2 for proteins \cite{Simon2025-it,Adams2025} and Evo2 for
  nucleic acids \cite{Brixi2025}). Since cLMs share the same foundations
  as these language models, SAEs offer a grounded way to answer the
  open question of \textit{what a chemical language model knows about
  molecules}.
  
  Concurrent with our work, \citet{cohen2025} and \citet{varadi2025}
  have also applied SAEs to understand cLM representations of molecules.
  While these studies focus on encoder-decoder and decoder-only
  architectures, respectively, we train SAEs on the residual stream of
  an encoder-only model, MolFormer-XL, and pursue a broader investigation
  into what the model has learned (discussed further in
  \cref{apx:related_work}). To answer our central question, we advance
  the analysis with four main contributions that together provide new
  insights into cLM molecular representations:
    \begin{enumerate}
      \item \textbf{InterMol.} We developed a web application for
        interactive visualization of SAE latent activations, available
        at \href{https://www.intermol.co/}{www.intermol.co}.
      \item \textbf{Molecular Grammar.} We identified SAE latents
        that activate on syntactic concepts, including branches,
        ring indices, and syntax parsing features besides
        atom-related features.
      \item \textbf{Learned Representation Shifts.} We show that
        position latents make non-canonical SMILES more disruptive
        to learned representations than invalid SMILES.
      \item \textbf{Interpretable Probing.} We identified
        pharmacologically relevant SAE latents from linear probe
        coefficients trained on downstream tasks.
    \end{enumerate}

\section{Background}
  \subsection{Molecular String Representations}
    Molecules are commonly represented as graphs, with atoms as nodes
    and bonds as edges. However, molecular graphs are computationally
    expensive for large sets of molecules. \citeauthor{Weininger1988-cg}
    (\citeyear{Weininger1988-cg}) tackled this limitation by introducing
    SMILES, Simplified Molecular Input Line Entry System, as a compact
    string representation. SMILES encodes molecular graphs by traversing
    atoms and bonds depth-first from any starting atom. Special
    characters indicate specific structural features, such as parentheses
    for branching and digits for ring closures.

    Although alternative molecular string representations have been
    developed, such as InChI \cite{Heller2013-gm}, DeepSMILES
    \cite{OBoyle2018-to}, and SELFIES \cite{Krenn2020-uf}, SMILES
    remains widely used due to its simplicity, human readability,
    widespread adoption across curated chemistry databases, and use as
    the input for many cLMs. Based on these advantages, we use SMILES
    as the main molecular string representation for our mechanistic
    interpretability study.

  \subsection{Chemical Language Models}
    cLMs apply natural language processing (NLP) techniques to learn
    from molecular string representations. Since the introduction of
    the transformer architecture, many cLMs have been developed to
    learn representations of chemical space through masked language
    modelling, including ChemBERTa variants \cite{chithrananda2020,ahmad2022}
    and MolFormer-XL \cite{Ross2022-jm}. Although ChemBERTa is widely
    used as a baseline, we selected MolFormer-XL for several reasons:
    (1) rotational position embeddings that allow efficient processing
    of long molecular strings; (2) pretraining on a large and diverse
    molecular corpus of 1.1 billion compounds from ZINC15 and PubChem
    compared to 77 million compounds from PubChem used for ChemBERTa;
    and (3) superior performance on most chemistry-related benchmarks.

    In particular, we investigation the open-weights variant of
    MolFormer-XL \cite{Ross2022-jm}. This variant was pretrained on
    only 10\% of the molecular SMILES used for the full model, yet
    achieved comparable performance on several chemistry benchmarks.
    Hereafter, we refer to this variant as MolFormer-10, which serves
    as the main model for our interpretability analysis.

\begin{figure*}[ht]
  \centering
  \includegraphics[width=\textwidth]{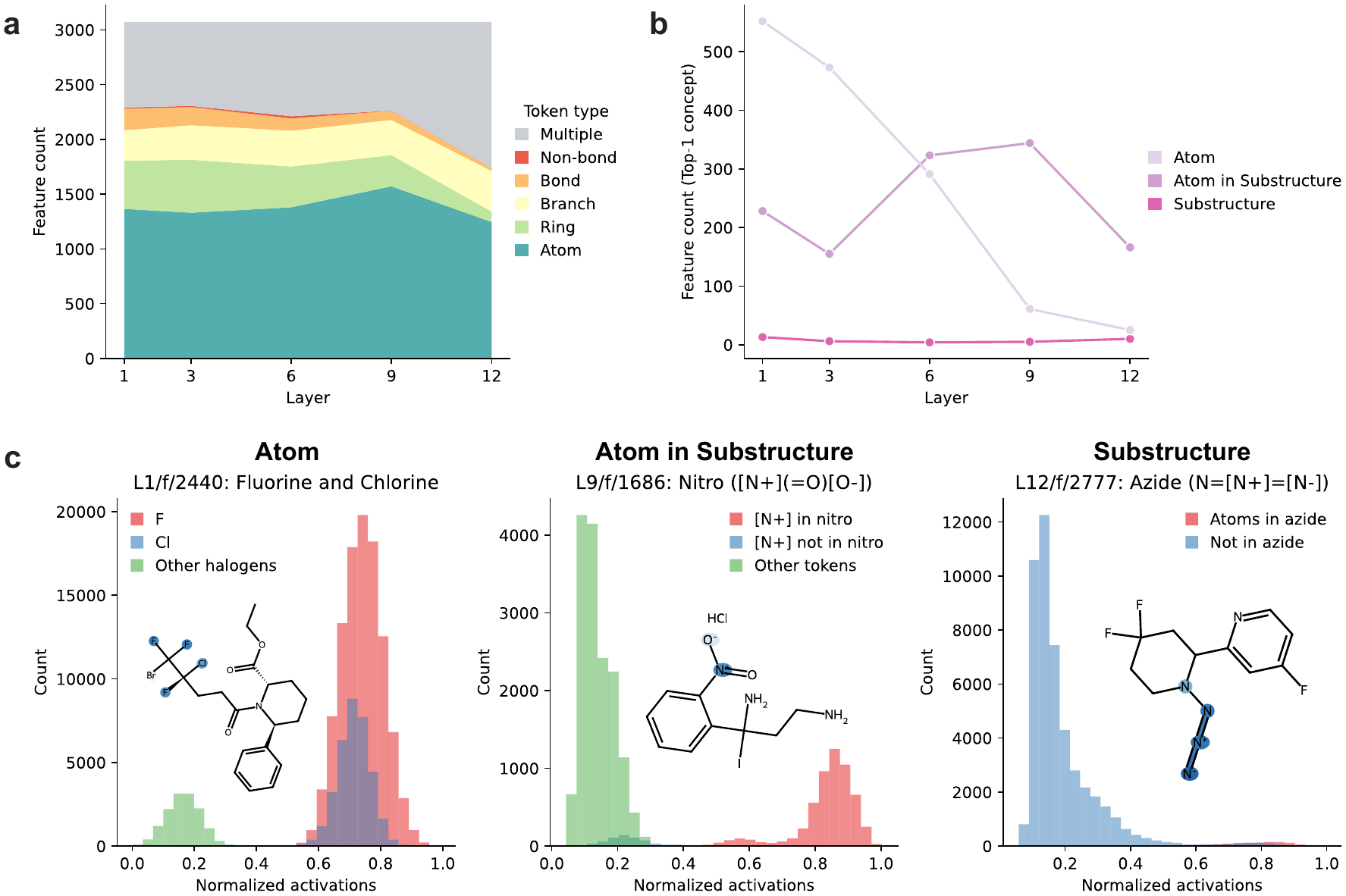}
  \caption{
    \textbf{Interpretable Molecular Features.} \textbf{(a)} SAE latents
    classified by dominant token type ($D \geq 0.6$) across layers.
    \textbf{(b)} Number of unique interpretable latents per layer
    matching their top-1 atom-related concept. \textbf{(c)} Examples
    of atom-related concepts encoded in latents across layers.
  }
  \label{fig:molecular_features}
\end{figure*}

\section{Methods}
  \subsection{Sparse Autoencoders}
    Sparse autoencoders (SAEs) are trained to reconstruct inputs while
    enforcing sparsity in the hidden layer. In this study, we use $k$-SAE
    \cite{gao2024}, which keeps only the top $k$ hidden-layer activations
    $z$ and nullifies the rest. The computation is expressed as:
      \[z = \text{TopK}(W_{enc}x + b_{enc})\]
      \[\hat{x} = W_{dec}z + b_{dec}\]
    where the encoder weights $W_{enc}$ project the residual stream
    into the SAE latent space and the decoder weights $W_{dec}$ reverse
    it. The reconstruction mean squared error
    \(\mathcal{L} = \lVert x - \hat{x} \rVert_2^2\) is used as the
    objective function to train the SAE.
  
  \subsection{Molecular Features}
    As part of understanding what the cLMs have learned from SMILES,
    we classified each SAE latent by its dominant token types and
    further map SAE activations to atom-related concepts: atom,
    atom-in-substructure, and substructure. For these concepts, we
    used SMARTS (SMILES Arbitrary Target Specification) notation, as
    SMARTS allows us to define specific chemical patterns that are
    easily interpretable. We provide the detailed steps in
    \cref{apx:token_types} (token types) and \cref{apx:atom_concepts}
    (atom-related concepts).

  \subsection{SMILES Variants}
    \textbf{Non-canonical SMILES} are variations that result from
    generating SMILES strings because molecules as graphs can be
    parsed at any starting atom or traversed in different ways.
    This can pose challenges for cLMs pretrained only on canonical
    SMILES, including MolFormer-10. To assess how MolFormer-10
    represents such variation, we built a toy dataset using RDKit
    \cite{Landrum2025-rj}.

    \textbf{Invalid SMILES} can appear as artifacts in generative
    model outputs or be accidentally introduced in databases (e.g.,
    mistakenly marking aromatic carbons \texttt{c} as non-aromatic
    ones \texttt{C}). As \citet{Skinnider2024-ks} suggests that
    invalid SMILES serve a beneficial role in generative models,
    understanding how cLMs internally represent them could inform
    the design of future cLMs. To this end, we adapted
    SMILES-corrupting scripts from \citet{Schoenmaker2023-ox} to
    build the toy dataset with further details in \cref{tab:errors}.

    We use RDKit-canonicalized SMILES throughout this paper unless
    otherwise specified.

  \subsection{Downstream Tasks}
    \textbf{ADMET}, or Absorption, Distribution, Metabolism, Excretion,
    and Toxicity, describes how drugs are processed by the body and is
    critical in ensuring that drugs are both safe and effective. We
    used ADMET prediction tasks, comprising 11 classification
    (\cref{tab:admet_cls}) and 10 regression tasks (\cref{tab:admet_reg}),
    to evaluate whether cLMs learned pharmacologically relevant features.
    The datasets were retrieved from the Therapeutics Data Commons
    (TDC) \cite{huang2021,Huang2022-hg}.

\section{Decomposing Molecular Understanding}
  \begin{figure*}[htb]
    \centering
    \includegraphics[width=\textwidth]{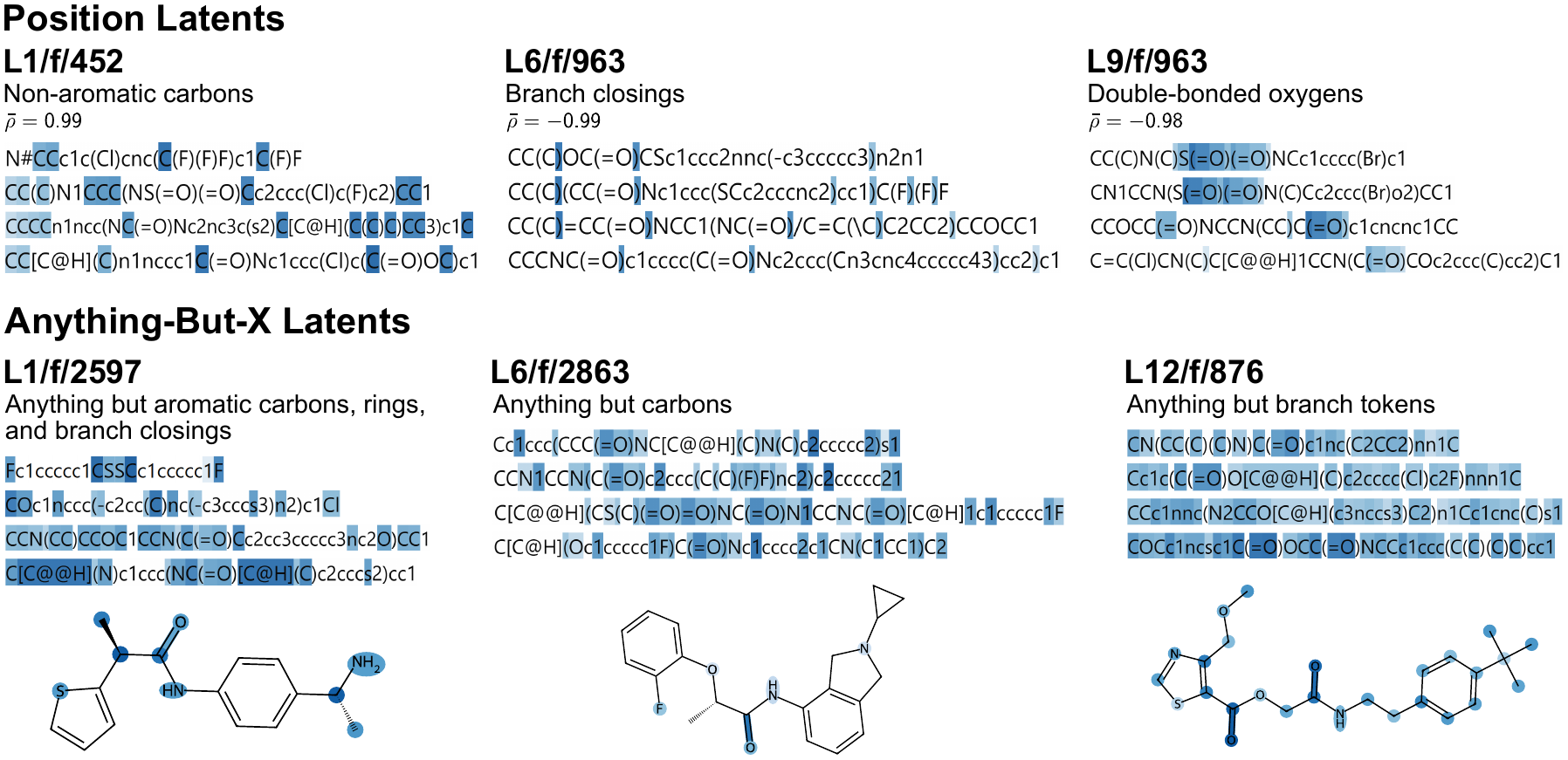}
    \caption{
      \textbf{Syntax Parsing Features.} \textbf{Position Latents}:
      \textbf{L1/f/452} shows increasing activations on token \texttt{C}
      toward the final token; \textbf{L6/f/963} shows decreasing
      activations on branch-closing token \texttt{)} toward the end;
      and \textbf{L9/f/1494} shows decreasing activations on the
      \texttt{(=O)} motif, possibly tracking double-bonded oxygens.
      \textbf{Anything-But-X Latents}: \textbf{L1/f/2597}
      ($p_{\text{zero}}=0.99$) does not activate on aromatic carbons
      ($p_{\text{ctx}}=1.0$), ring indices ($p_{\text{ctx}}=0.99$),
      and branch closings ($p_{\text{ctx}}=0.99$); \textbf{L6/f/2863}
      ($p_{\text{zero}}=0.99$) does not activate on any carbons,
      including \texttt{C} ($p_{\text{ctx}}=0.99$), \texttt{c}
      ($p_{\text{ctx}}=0.99$), \texttt{[C@H]} ($p_{\text{ctx}}=0.99$),
      \texttt{[C@@H]} ($p_{\text{ctx}}=0.99$), \texttt{[C@]}
      ($p_{\text{ctx}}=0.95$), and \texttt{[C@@]} ($p_{\text{ctx}}=0.99$);
      and \textbf{L12/f/876} ($p_{\text{zero}}=0.98$) does not
      activate on branch openings ($p_{\text{ctx}}=0.99$) and closings
      ($p_{\text{ctx}}=0.99$).
    }
    \label{fig:syntax_features}
  \end{figure*}

  Uncovering chemically meaningful features from SAE latents is
  challenging due to the diverse token types in SMILES. SMILES consists
  of five main token types: atoms, ring indices that number rings,
  branches marking branching points, bond notations representing
  interatomic connectivity (e.g., \texttt{=} for double bonds and
  \texttt{\#} for triple bonds), and disconnections or non-bonds
  denoting molecular fragments.
  
  Examining the token-type distribution of each latent across
  MolFormer-10 layers via Simpson's dominance index $D$ reveals
  that syntax-related token types are least frequent across layers,
  while nearly half of the latents up to Layer 9 activate on
  atom-type tokens (\hyperref[fig:molecular_features]{Figure 1a}).
  This is because syntax-related tokens---rings, branches, bonds,
  and non-bonds---contain fewer unique tokens by their nature
  as molecular grammar rather than atomic composition. In the last
  layer, however, latents activating on multiple token types become
  as common as those activating on atom types, signifying more
  diffuse activation patterns compared to the token-specific
  patterns observed in early and middle layers.

  \subsection{Atom-Related Concepts}
    We next match atom-type-preferring latents against a predefined
    library of atom-related concepts (\cref{apx:atom_concepts}).
    By selecting the concept with the highest F1 score
    ($\text{F1} \geq 0.5$) for each latent, we measure the number
    of interpretable latents in \hyperref[fig:molecular_features]
    {Figure 1b}, with several examples shown in
    \hyperref[fig:molecular_features]{Figure 1c}.

    Across layers, latents matching atom-level concepts show a
    bottom-up trend, from atom-focused activation to atom-in-context
    and substructure activation patterns
    (\hyperref[fig:molecular_features]{Figure 1b}). Atom-in-context
    latents suggest a more efficient SAE approach to decomposing
    molecules based on common structural contexts, especially
    in Layers 6 and 9. For example, rather than learning separate
    latents for carboxylic acids (\texttt{C(=O)O}) and amides
    (\texttt{C(=O)N}), the model appears to learn shared atom-pair
    latents for their common motifs: \texttt{C=O}, \texttt{C-N},
    and \texttt{C-O}. When an amide is present, the first two
    latents activate together; when a carboxylic acid is present,
    the first and third activate. This decomposition may enable
    the cLM to reuse these latents across more complex molecular
    structures.
    
    Substructure-level concepts were expected to enrich in later
    layers, but instead show an unclear trend as seen in
    \hyperref[fig:molecular_features]{Figure 1b}. Nonetheless,
    given that most of these latents activate across multiple
    token types (\hyperref[fig:molecular_features]{Figure 1a}),
    we argue that this indicates the cLM may not learn all
    structural contexts, but rather syntactic ones of SMILES.

  \subsection{Syntax Parsing Features}
    We also discover latents whose activation patterns extend
    beyond atom-related concepts. We term these syntax parsing
    features, arguing that they may provide contextual information
    enabling the cLM to process inputs and build internal
    molecular representations. We identify two types: position
    latents and anything-but-X latents.

    \subsubsection{Position Latents}
      A recent study \cite{sun2025} found that several dense
      latents (i.e., those activating on multiple tokens) have
      activations correlated with relative token positions within
      specific contexts. We identify similar latents in our SAE,
      which are more prominent in early layers and decrease
      toward later layers \cref{fig:position_latents}, consistent
      with early layers processing the tokenized SMILES input. As
      shown in \cref{fig:syntax_features}, L1/f/452 and L6/f/963
      strongly activate on later non-aromatic carbon tokens and
      earlier branch-closing tokens, respectively, and decrease
      toward the opposite end with no preference of specific
      context. Unlike simple position tracking, L9/f/1494 tracks
      double-bonded oxygens within a branch from opening to closing,
      indicating syntactic substructure completeness.
    
    \subsubsection{Anything-But-X Latents}
      \label{main:anything_but_x}

      Anything-but-X latents activate on all tokens except
      specific tokens or contexts. Previously, \citet{fel2025}
      identified similar latents, termed "elsewhere" latents,
      in a vision transformer. There are two points that make
      anything-but-X latents differ from "elsewhere" latents:
      (1) "elsewhere" latents activation is context-dependent,
      whereas ours activate independently; and (2) consequently,
      "elsewhere" latents imply learned context negations, while
      the function of anything-but-X latents remains
      unclear---possibly an artifact of the small SAE size.

      We characterize these latents using two proportions
      that measure latents exclusivity: (1) the ratio of
      non-activated tokens within a specific context to all
      \textbf{non-activated tokens} ($p_{\text{zero}}$), and
      (2) the ratio of non-activated tokens within a specific
      context to all \textbf{tokens with that context in the
      dataset} ($p_{\text{ctx}}$). As an example
      (\cref{fig:syntax_features}), L6/f/2863 exhibits this
      characteristic by not activating on common carbon-related
      tokens, suggesting it may help discriminate heteroatoms.
      While this analysis serves as a preliminary finding,
      a more thorough analysis using larger SAE dimensions
      and circuit-level analysis of the cLM residual stream
      is needed to confirm the exclusivity of anything-but-X
      latents and rule out polysemanticity.

  \section{SMILES Variants Perturb Representations}
    \label{main:alt_repr}

    CLMs that learn molecular structures should ideally be
    invariant to variations in molecular string notation.
    To test this, we analyze how cLM internal representations
    encode SMILES variants by examining the similarity of
    max-pooled SAE latents (SAE embeddings) and assessing
    how individual latents respond to shifts in internal
    molecular representation.

\begin{figure*}[ht]
      \centering
      \includegraphics[width=\textwidth]{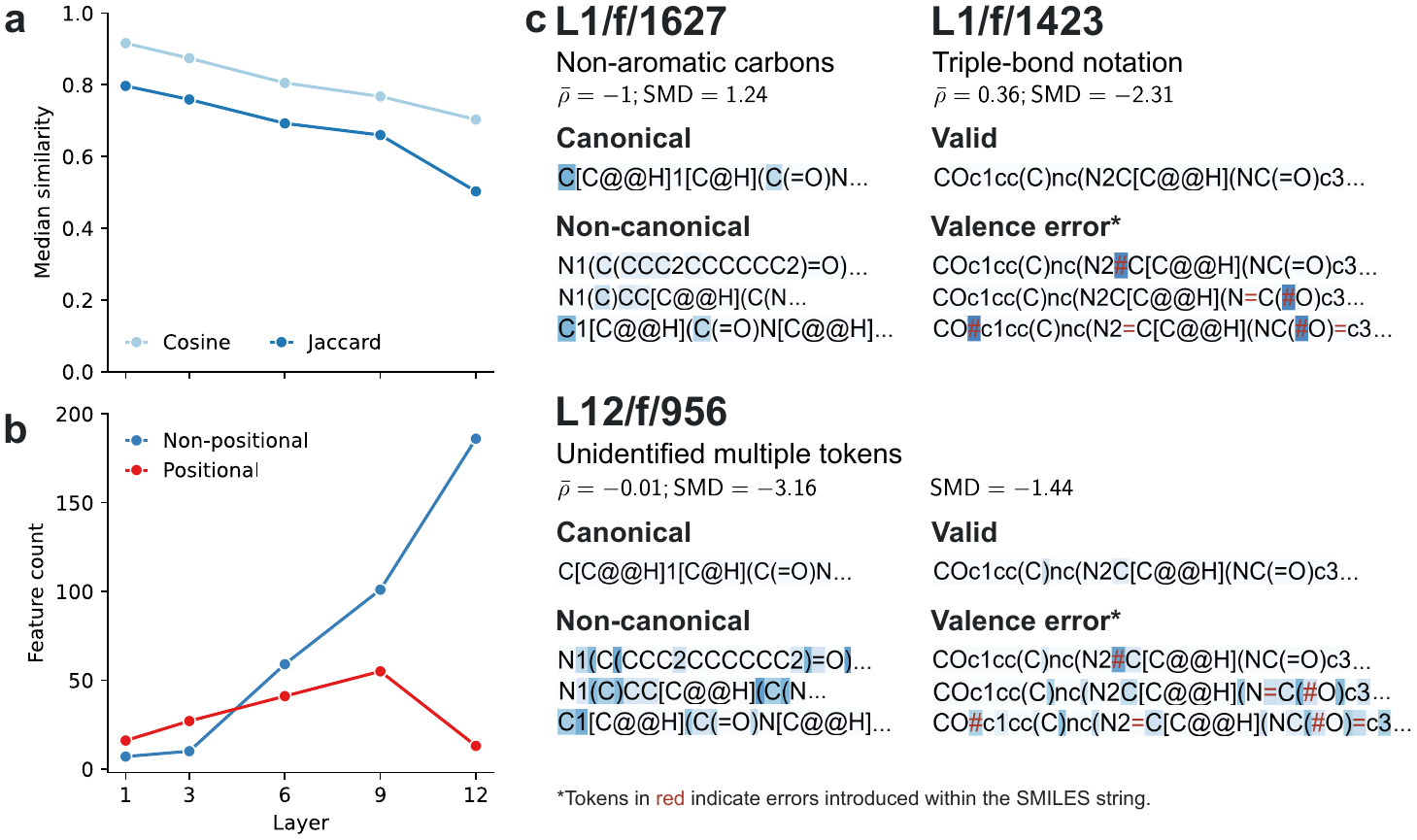}
      \caption{
        \textbf{Position Latents Shift Learned Representations.}
        \textbf{(a)} Median cosine and Jaccard similarities between
        max-pooled SAE activations of canonical and non-canonical
        SMILES pairs. \textbf{(b)} Number of SAE latents with 
        $|\text{SMD}| \geq 0.8$ classified as non-positional and
        positional across layers for the augmented SMILES. \textbf{(c)}
        Examples of features with significant SMD in non-canonical
        and valence error cases.
      }
      \label{fig:alt_repr}
    \end{figure*}

\begin{figure}[H]
  \centering
  \includegraphics[width=\columnwidth]{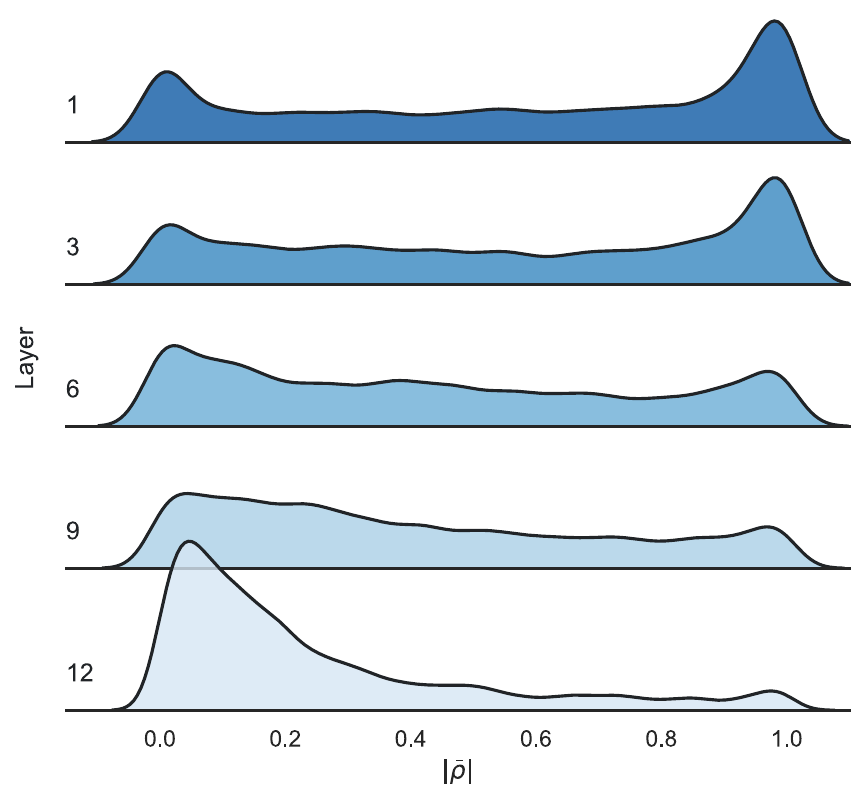}
  \caption{
    \textbf{Position Latents Decrease Toward Later Layers.}
    Distribution of absolute Fisher-averaged Spearman coefficients
    across layers. Peaks near $|\bar{\rho}|\approx1$ at early
    layers indicate a significant presence of position latents,
    while later layers show a flattened distribution skewed
    toward $|\bar{\rho}|\approx0$, suggesting that fewer position
    latents are present and contribute less positional information
    to the layer's internal representation.
  }
  \label{fig:position_latents}
\end{figure}

    For SAE representation similarity, we compute cosine
    similarity between raw values and Jaccard similarity
    between binarized (active/inactive) values across
    canonical and variant SMILES pairs. For the latent-specific
    analysis, we compute the standardized mean difference
    (SMD) as an effect size for each latent $f$.
      \[\text{SMD}_f = \frac{\mu_f^c - \mu_f^{nc}}{\sigma_{\text{pooled}}}\]
    where $\mu_f^c$ and $\mu_f^{nc}$ are the mean max-pooled
    activations of latent $f$ over the reference (canonical
    or valid) and comparison (non-canonical or invalid)
    groups, respectively, and $\sigma_{\text{pooled}}$ is
    the pooled standard deviation across both groups. Results
    of both analyses for the augmented SMILES are presented
    in \hyperref[fig:alt_repr]{Figure 4a} and
    \hyperref[fig:alt_repr]{Figure 4b}, respectively.

    \subsection{Augmented SMILES}
      As illustrated in \hyperref[fig:alt_repr]{Figure 4a},
      the median cosine similarity of SAE embeddings between
      canonical and non-canonical SMILES is higher than
      the median Jaccard similarity across layers, with
      the largest discrepancy at layer 12. This indicates
      that while the set of active latents may differ between
      canonical and non-canonical SMILES, those that activate
      in both share similar activation magnitudes, suggesting
      they likely capture structural rather than purely
      syntactic patterns.

      Latents with large effect size ($|\text{SMD}| \geq 0.8$)
      are mostly position latents in early layers
      (\hyperref[fig:alt_repr]{Figure 4b}). For example,
      as shown in \hyperref[fig:alt_repr]{Figure 4c},
      non-canonical SMILES suppress the max-pooled activations
      of L1/f/1627 when the first token changes from
      \texttt{C} to \texttt{N}. From layer 6 onward,
      non-position latents with large $|\text{SMD}|$
      that activate more strongly in non-canonical SMILES
      predominate (\cref{fig:suppl_nc}), suggesting that
      early-layer position latents disruption may propagate
      into deeper layers.

    \subsection{Corrupted SMILES}
      Of the five error types, syntax errors exhibit the
      widest similarities distribution (\cref{fig:suppl_sim}),
      likely because they introduce more disruptive
      modifications, such as SMILES truncation, compared
      to simpler token insertion or duplication in other
      error types (see \cref{tab:errors}). Interestingly,
      aromaticity and kekulization errors have the highest
      median similarities across layers, despite mainly
      swapping atoms that contribute differently to the
      ring's $\pi$-electron system. This suggests that
      the SAE recognizes whether an atom belongs to an
      aromatic ring, but is insensitive to the specific
      atomic identity within that ring.

      Regarding latent-specific differences, almost all
      error types affect few latents significantly, except
      for valence errors (\cref{fig:suppl_err}). This is
      because valence errors introduce new bond tokens
      that activate related latents, resulting in high-SMD
      latents that are primarily non-positional. Notably,
      we identified a latent in layer 12, L12/f/956, that
      consistently shows a large effect size across all
      SMILES variants, with stronger activation when the
      input is augmented (\cref{fig:suppl_nc}) or corrupted
      (\cref{fig:suppl_err}), implying that this latent
      may function as a canonicality and validity detector.

\section{Chemically Relevant Latents}
  We used ADMET tasks to search for chemically relevant latents.
  Following \citet{Adams2025}, we employ SAE latents as input
  features to a simple linear model, whose coefficients directly
  reflect the contribution of each latent. The InterMol visualizer
  is used to interpret the top-contributing latents and identify
  their physicochemical-related features.

  \subsection{Performance on ADMET Tasks}
    Overall, SAE embeddings perform better on 16 out of 21
    ADMET tasks, followed by MolFormer-10 (4 tasks), and
    physicochemical descriptors (1 task). SAE latents likely
    capture chemical features similar to the baselines,
    such as atom-in-substructure concepts analogous to the
    atomic environments encoded in ECFPs (\cref{fig:suppl_ecfp2}),
    while potentially encoding more complex features not
    explicitly captured by the baselines. We examine 4
    tasks for further pharmacological interpretation in
    \cref{submain:interpreting_probes}.

  \begin{figure*}[ht]
    \centering
    \includegraphics[width=\textwidth]{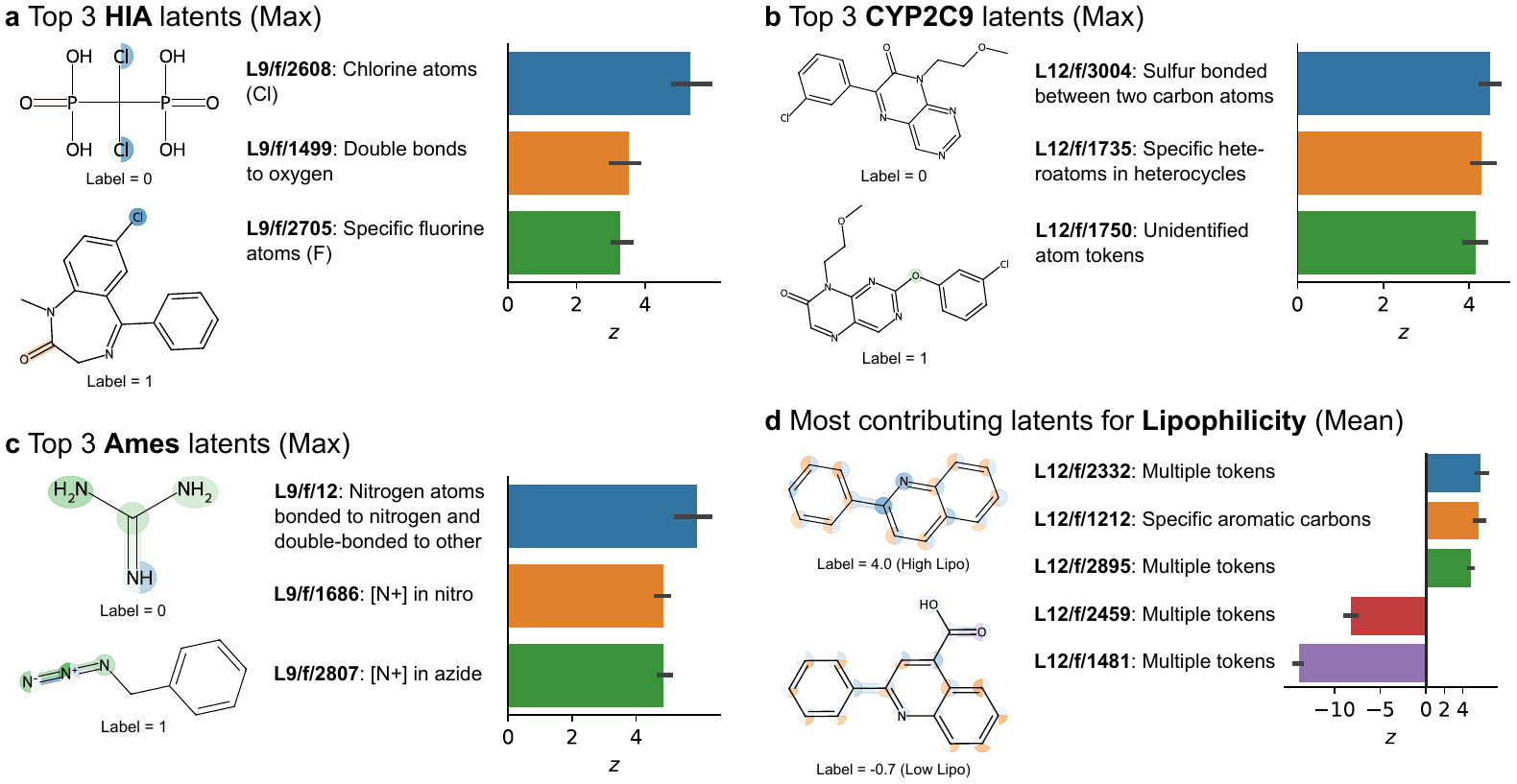}
    \caption{
      \textbf{Linear Probe Identifies Pharmacologically
      Relevant SAE Latents.} We show the top 3 latents
      by standardized linear probe coefficient, $z(\beta_{f})$,
      averaged across triplicates, for \textbf{(a)} human
      intestinal absorption (HIA), \textbf{(b)} CYP2C9
      inhibition, and \textbf{(c)} Ames mutagenicity
      classification; and \textbf{(d)} top 3 positive
      and top 2 negative latents for the lipophilicity
      regression task.
    }
    \label{fig:admet_interp}
  \end{figure*}
  
  \subsection{Interpreting Probes}
    \label{submain:interpreting_probes}
    
    \textbf{Human Intestinal Absorption} (HIA) evaluates the ability
    of drug molecules to be absorbed into the bloodstream from the
    gastrointestinal (GI) tract \cite{Hou2007-rb}. Our linear probe
    identifies three latents corresponding to two structural motifs
    associated with absorptivity (\hyperref[fig:admet_interp]
    {Figure 6a}). The first is halogen substituents, such as chloro-
    (L9/f/2608) and fluoro- (L9/f/2705), both of which are known
    to improve lipophilicity and thus GI absorptivity
    \cite{Dudek2025-bj}. The second is double-bonded oxygen;
    L9/f/1499 specifically fires on the double bond token. Although
    this motif increases solubility, a balance between solubility
    and lipophilicity is needed for optimal GI absorption---making
    this latent meaningful in predicting HIA.

    \textbf{CYP2C9 Inhibition.} CYP2C9 is a cytochrome P450
    isoenzyme whose inhibition hinders the body's ability to
    metabolize co-administered drugs, potentially leading to
    toxicity \cite{Veith2009-kg}. Among the three SAE latents
    identified, only one is pharmacologically relevant:
    L12/f/1735, which activates on heteroatoms in heterocycles,
    aligned with common pharmacophoric features of CYP2C9
    inhibitors reported in \citet{Beck2021-iu}. However,
    other pharmacophoric motifs absent from the top-3 latents,
    particularly relevant for inhibitory tasks, may reside
    in other SAE latents as supported by \cref{fig:suppl_taskablate}.
    While L12/f/1750 has no clear interpretation, its activation
    on the oxygen atom between aromatic rings
    (\hyperref[fig:admet_interp]{Figure 6b}) is able to
    differentiate between two molecules with similar structures
    but different inhibitory activity.

    \textbf{Ames Mutagenicity.} This refers to the ability of
    a molecule to induce DNA damage and cause genetic alterations.
    The dataset is constructed from the Ames bacterial test,
    the most widely used \textit{in vitro} assay for assessing
    mutagenic potential \cite{Xu2012-sv}. We found three latents
    corresponding to mutagenic properties. The top latent,
    L9/f/12, activates on nitrogens bonded to other nitrogens,
    commonly found in mutagenic motifs such as the N-N=O of
    nitrosamines. The remaining two latents both activate more
    strongly on \texttt{[N+]}: L9/f/1686 on nitro groups and
    L9/f/2807 on azides (the latter seen in
    \hyperref[fig:admet_interp] {Figure 6c}), both well-known
    mutagenic motifs \cite{Nepali2019-xg,Gruz2022-re}.

    \textbf{Lipophilicity} measures a molecule's relative
    tendency to dissolve in lipids over water, directly
    influencing its ability to be absorbed in the GI tract.
    As shown in \hyperref[fig:admet_interp]{Figure 6d}, no
    latents are found to activate on specific concepts;
    however, the activation of L12/f/1212 on aromatic carbons
    provides an interpretable basis for increased lipophilicity,
    as aromatic carbons are known to be non-polar. In contrast,
    for the low lipophilicity sample, the additional functional
    motif -COOH is activated with L12/f/1481, which negatively
    influences the predicted lipophilicity, consistent with
    the lipophobic nature of the motif.

\section{Discussion}
  In this work, we presented a mechanistic analysis of an
  encoder-only cLM, MolFormer, using SAEs. We found that
  molecular representations are built hierarchically,
  progressing from atom-level concepts at early layers
  toward more nuanced atom-in-substructure and substructure-level
  concepts at later layers. This hierarchical organisation
  is further reflected in the performance on ADMET tasks,
  where most results exceed the baseline methods of molecular
  fingerprints and descriptors, showing that MolFormer
  encodes pharmacologically relevant molecular features.
  To support the further exploration of many remaining
  uninterpreted latents, we introduce InterMol, an
  open-source visualizer for that may reveal new insights
  into the inner workings of cLMs.
  
  In addition to atom-related features, MolFormer also has its own
  mechanistic way of parsing the molecular grammar encoded in
  SMILES input, using prevalent position latents to track atom-type
  token positions and syntactic substructure completeness. Yet,
  this mechanism may introduce an invariance challenge for cLMs
  pretrained only on canonical SMILES, as it may cause
  position-dependent disruptions to learned representations
  across different SMILES notations of the same molecule. Therefore,
  a good training design, like exposing the cLM to augmented
  SMILES or, with incorporating contrastive learning, may help
  address this issue by reducing positional bias, as supported
  by recent studies \cite{bjerrum2017,D5DD00028A}. On the contrary,
  invalid SMILES produce more subtle shifts in learned
  representations. This may also partially explain the behavior
  of autoregressive cLMs that is occasionally generate invalid SMILES
  during molecular generation, though additional study is needed
  to confirm this.
  
  \textbf{Limitations.} Despite these findings, our work
  has several limitations. First, we focus only on MolFormer,
  which may not generalize to other encoder-only cLMs
  due to differences in architecture, SMILES input, or
  atom-wise tokenization, though our findings provide a
  foundation for future cLM designs, with MolFormer itself
  already aligned with recent recommendations of
  \cite{Fender2025-rq}. Second, although we randomly sampled
  from available molecular SMILES to reduce structural
  biases, our SAE training dataset remains limited in scale.
  Third, MolFormer uses rotary position embeddings (RoPE)
  in each attention head computation, which entangles
  relative positional information and molecular language
  signals, making attention-based circuit analysis difficult
  to apply. We therefore limit our analysis to the molecular
  features encoded in the residual stream of MolFormer.
  Finally, interpreting SAE latents remains challenging
  given the lack of gold-standard molecular concepts, unlike
  those available for proteins via UniProt and DNA via
  GenBank, and we therefore consider some of our
  interpretations of complex activation patterns as
  preliminary---we invite the community to refine or correct
  them through InterMol.
  
  \textbf{Future Work.} Our findings suggest several
  promising directions for future research. One direction
  involves improving the training design of cLMs. While
  SMILES augmentation may help reduce positional bias as
  previously discussed, multimodal representation learning
  offers a more grounded approach. For instance, joint-embedding
  predictive architecture (JEPA) could be incorporated by
  integrating both molecular strings and structures,
  following recent evidence of successful LLM-JEPA integration
  \cite{huang2025}. This multimodality may help cLMs learn
  richer molecular representations, less dominated by
  syntax-related features. Another intriguing direction is
  deepening the mechanistic analysis of cLMs by employing
  cross-layer transcoders (CLTs) to identify more causally
  grounded molecular cirucits, recently applied to understand
  protein circuits in protein language models \cite{tsui2026}
  and enable downstream applications like steering autoregressive
  cLMs for self-directed molecular generation.

\section*{Acknowledgements}
    We declare no conflicts of interest.

\section*{Impact Statement}
    This work advances our understanding of cLMs by revealing
    interpretable molecular features encoded in their representations.
    Our findings may support rational molecular design and
    accelerate drug discovery. We do not foresee ethical risks
    arising from this research.

\bibliography{example_paper}
\bibliographystyle{icml2026}

\newpage
\appendix
\onecolumn
\section*{Appendix}
\section{Code Availability}
  The code to reproduce the experiments is available at
  \url{https://github.com/ckennetha/intermol}.

\section{Related Work}
  \label{apx:related_work}

  \textbf{Mechanistic Interpretability of Chemical Language Models.}
  Understanding how cLMs work internally may help in designing
  better cLMs that enable the generation of \textit{de novo}
  molecules with desired properties. In parallel with our work,
  \citet{cohen2025} and \citet{varadi2025} have explored this
  by applying SAEs to an encoder-decoder model (SMI-TED) and
  and a decoder-only model, respectively. Both studies find that
  SAE latents correspond to chemically meaningful concepts,
  including substructure motifs and physicochemical properties.
  \citet{cohen2025} also demonstrates pharmacological-level
  abstractions through features that group structurally
  dissimilar compounds sharing certain functionality. Both
  works also show controllable generation of specifc molecules
  through SAE-based interventions. Additionally, \citet{varadi2025}
  extend their analysis to circuit-level mechanisms, identifying
  attention heads that track ring and branch completeness and
  a linear representation of valence capacity in the residual
  stream that modulates bond-order predictions. Despite these
  contributions, existing studies have not yet examined encoder-only
  models, which are widely used for predictive tasks. Furthermore,
  the global role of specific latents in capturing syntactic
  variation across SMILES representations has not yet been
  elucidated, which is integral to the development of cLMs.

\section{Experimental Details}
  \subsection{Sparse Autoencoders}
    \textbf{Training Setup.} We trained the SAEs on the 768-dimensional
    residual streams of MolFormer-10 layers: 1 (early), 3, 6, 9,
    and 12 (final), with the SAE latent dimension of 3,072 (4x
    expansion factor) and $k=128$. For the training dataset, we
    used approximately 2.2 million SMILES randomly sampled from
    ZINC15 \cite{Irwin2005-dd} and PubChem \cite{Kim2019-id}, the
    same data pool used to train MolFormer-XL.

    \textbf{Normalization.} For ease of interpretation, we
    normalized SAE latents using $\sim$250,000 molecular SMILES
    randomly sampled from the SAE training dataset. Following
    \citet{Simon2025-it}, for each latent in each SAE, we used
    its maximum activation to rescale the encoder and decoder
    weights, resulting the activations to the range [0, 1] while
    preserving reconstruction. However, there might be some edge
    cases in which the activations exceed 1.
  
  \subsection{Feature Discovery in the cLM}
    \subsubsection{Datasets}
      \label{apx:feature_dataset}
      The dataset for the feature analysis was constructed by
      randomly sampling another 500,000 SMILES from the same data
      pool that are non-overlapping with the training dataset.
      We refer to this as evaluation set. Then, we randomly split
      the evaluation set in half into validation and test sets.

      In SMILES variation experiments, we randomly picked 20,000
      SMILES from the validation set. For the augmented toy dataset,
      we generated up to 5 non-canonical SMILES per canonical
      SMILES, resulting in around 100,000 canonical/non-canonical
      pairs. For the perturbed toy dataset, we generated all
      possible combinations of error counts (1, 2, 4) and five
      error types: syntax, parentheses (unmatched barckets),
      rings (unclosed rings), aromaticity (non-ring atoms marked
      as aromatic and kekulization errors), and valence (atoms
      exceeding the maximum number of bonds), as shown in
      \cref{tab:errors}. This yields around 300,000 valid/invalid
      pairs.

      \begin{tabularx}{\textwidth}{lXl}
    \caption{
        \textbf{Invalid SMILES Generation.} Error variations are randomly
        introduced for each error type with examples of resulting invalid
        SMILES. Red tokens on the left indicate deletions, while those on
        the right indicate substitutions or insertions.
    }
    \label{tab:errors} \\
    \toprule
        Error type & Variation & Invalid SMILES \\
    \midrule
    \endfirsthead

    \toprule
        Error type & Variation & Invalid SMILES \\
    \midrule
    \endhead

    \midrule
    \endfoot

    \bottomrule
    \endlastfoot
        \multirow{5}[-20]{*}{Rings} & Remove a ring index & \texttt{c\textcolor{red}{1}ccccc1 $\rightarrow$ cccccc1} \\
            & Replace a ring index by $\pm1$ & \texttt{c1ccccc1 $\rightarrow$ c\textcolor{red}{2}ccccc1} \\
            & Replace a ring index with different existing ring index & \texttt{C1C2CC12 $\rightarrow$ C\textcolor{red}{2}C2CC12} \\
            & Duplicate a ring opening index & \texttt{c1ccccc1 $\rightarrow$ c1\textcolor{red}{1}ccccc1} \\
            & Replace a random token with a ring index & \texttt{c1ccccc1 $\rightarrow$ c1cc\textcolor{red}{1}cc1} \\
        \multirow{5}[-20]{*}{Parentheses} & Insert a random \texttt{(} or \texttt{)} token & \texttt{CC(C)O $\rightarrow$ C\textcolor{red}{(}C(C)O} \\
            & Remove a random branch token & \texttt{CC\textcolor{red}{(}C)O $\rightarrow$ CCC)O} \\
            & Switch a pair of parentheses & \texttt{CC(C)O $\rightarrow$ CC\textcolor{red}{)}C\textcolor{red}{(}O} \\
            & Switch a branch opening token \texttt{(} to \texttt{)} & \texttt{CC(C)O $\rightarrow$ CC\textcolor{red}{)}C)O} \\
            & Switch a branch closing token \texttt{)} to \texttt{(} & \texttt{CC(C)O $\rightarrow$ CC(C\textcolor{red}{(}O} \\
        \multirow{5}[-20]{*}{Aromaticity} & Convert a non-aromatic to aromatic & \texttt{c1ccoc1 $\rightarrow$ c1cco\textcolor{red}{C}1} \\
            & Convert an aromatic atom with $\pi$ [\texttt{c}, \texttt{n}] to $2\pi$ electron & \texttt{c1ccoc1 $\rightarrow$ c1cco\textcolor{red}{s}1} \\
            & Insert a random \texttt{c} or \texttt{n} token into aromatic ring & \texttt{c1ccoc1 $\rightarrow$ c1cco\textcolor{red}{n}c1} \\
            & Convert an aromatic atom with $2\pi$ [\texttt{[nH]}, \texttt{o}, \texttt{s}] to $\pi$ electron & \texttt{c1ccoc1 $\rightarrow$ c1cc\textcolor{red}{c}c1} \\
            & Remove an aromatic atom with $\pi$ electron & \texttt{c1cco\textcolor{red}{c}1 $\rightarrow$ c1cco1} \\
        \multirow{7}[-28]{*}{Syntax} & Insert a random bond [\texttt{-}, \texttt{=}, \texttt{\#}] or branch [\texttt{(}, \texttt{)}] token at start & \texttt{CC(=O)CO $\rightarrow$ \textcolor{red}{=}CC(=O)CO} \\
            & Insert a random bond or branch token at end & \texttt{CC(=O)CO $\rightarrow$ CC(=O)CO\textcolor{red}{(}} \\
            & Insert a random bond next to an existing bond token & \texttt{CC(=O)CO $\rightarrow$ CC(\textcolor{red}{\#}=O)CO} \\
            & Insert a random bond token before a ring opening & \texttt{c1ccccc1 $\rightarrow$ c\textcolor{red}{=}1ccccc1} \\
            & Insert a random bond token before branch opening & \texttt{CC(=O)CO $\rightarrow$ CC\textcolor{red}{-}(=O)CO} \\
            & Duplicate a branch opening token & \texttt{CC(=O)CO $\rightarrow$ CC\textcolor{red}{(}(=O)CO} \\
            & Remove tokens between \texttt{(} and \texttt{)} and optionally insert a random bond or branch token & \texttt{CC(\textcolor{red}{=O})CO $\rightarrow$ CC()CO} \\
        Valence & Increase the bond order of 1 to 2 or 3 and 2 to 3 & \texttt{CC(=O)CO $\rightarrow$ C\textcolor{red}{=}C(=O)CO} \\
\end{tabularx}

    \subsubsection{Dominant Token Type Classification}
      \label{apx:token_types}

      MolFormer tokenizes SMILES input using atom-wise tokenization
      \cite{Ross2022-jm}, which produces tokens that can be
      classified further into separate token types. To characterize
      which token types each latent preferentially activates on,
      we gather all activated tokens across the validation set
      and grouped them by type using regex pattern matching. Tokens
      matching \verb|[\(\)]| are classified as branch type,
      \verb|[\-\=\#\$\\\/]| as bond type, \verb!(\%[0-9]{2}|[0-9])!
      as ring type, \verb|.| as non-bond type, and the rest
      as atom type.

      \textbf{Metric.} We use Simpson's dominance index $D$,
      which is the complement of Simpson's diversity index $1-D$,
      using the following formula.
        \[D = 1 - \sum_{i=1}^{n_t} p_i^2\]
      where $p_i$ is the proportion of token type $i$ across $n_t$
      total unique token types for a given latent $f$. Then, we
      assign the token type with the highest proportion as the
      dominant token type if $D \geq 0.6$; otherwise, we label
      it as "Multiple", indicating it activates on multiple
      token types.

    \subsubsection{Atom-Related Concept Association}
      \label{apx:atom_concepts}

      Atom-related concepts are defined using SMARTS patterns
      from two sources: predefined SMARTS libraries from the
      \hyperlink{https://github.com/rdkit/rdkit/tree/master/Data}
      {RDKit GitHub repository} and predominant atom-in-substructure
      SMARTS discovered in the validation set. The former
      comprises 533 patterns related to common functional motifs,
      which were later expanded to 2,473 unique patterns by
      re-enumerating each atom within every substructure. The
      latter were generated for each atom by incorporating its
      neighboring atoms and connectivity, deduplicated for the
      same-substructure SMARTS, and selected with occurrence
      in more than 10 molecules, totaling 2,483 unique patterns.
      Each of these patterns was then grouped into three
      categories: \textbf{atom} (single atom; e.g., \texttt{Cl}
      for chlorine atoms), \textbf{atom-in-substructure} (single
      atom within a substructure; e.g., \texttt{[\$([N+])](=O)[O-]}
      for cationic nitrogen in nitro group), and \textbf{substructure}
      (group of connected atoms; e.g., \texttt{N=[N+]=[N-]} for
      azides).

      \textbf{Filtering SAE Latents.} Before evaluating the
      association between atom-related concepts (i.e., pattern)
      and SAE latents, we employed latent filtering by selecting
      the top 64 latents using the standardized mean difference
      (SMD) between the activations on atom tokens on which the
      pattern is present, $wc$, and the activations on atom
      tokens on which the pattern is not present, $oc$, computed
      on 20\% of the validation set. For each latent $f$, we compute:
        \[\text{SMD}_f = \frac{\mu_f^{wc} - \mu_f^{oc}}{\sigma_{\text{pooled}}}\]
      where $\mu_f^{wc}$ and $\mu_f^{oc}$ are the mean activations
      of latent $f$ and $\sigma_{\text{pooled}}$ is the pooled
      standard deviation from both groups.

      \textbf{Evaluating Associations.} Following the work of
      \citet{Simon2025-it}, we binarized normalized SAE
      activations---latent-on/latent-off---using several activation
      thresholds of 0, 0.1, 0.2, 0.35, 0.5, 0.6, and 0.8. For each
      pattern, we first evaluated all possible latent-pattern pairs
      on the validation set and selected those with F1 score
      $\geq 0.5$ to reduce the number of tested patterns. Then,
      we re-evaluated these pairs on the test set and reported
      those passing the same criterion. The identified associations
      can be explored in our interactive visualizer, InterMol.

      \textbf{Metrics.} For each latent-pattern-threshold triplet,
      the association was evaluated using precision, recall, and
      F1 score, with modified recall and F1 score for substructural
      patterns. The metrics are defined as follows.
        \[\text{Pr} = \frac{TP}{TP + FP} \qquad \text{Re} = \frac{TP}{TP + FN} \qquad \text{F1 score} = \frac{2 \times \text{Pr} \times \text{Re}}{\text{Pr} + \text{Re}}\]
        \[\text{Re}_{sub} = \frac{TP_{sub}}{n_{sub}} \qquad \text{F1 score}_{sub} = \frac{2 \times \text{Pr} \times \text{Re}_{sub}}{\text{Pr} + \text{Re}_{sub}}\]
      where $TP_{sub}$ is the number of substructures where the
      latent activates on all atoms and $n_{sub}$ is the total number
      of substructures present in the dataset.

    \subsubsection{Syntax Parsing Features}
      \textbf{Position Latents.} In identifying position latents,
      we used Spearman correlation coefficient, $\rho$, between
      the activations and the relative token position within a
      SMILES string. We set a minimum of 3 activated tokens to
      calculate the correlation, allowing for rarely-occurring
      syntactic tokens, such as bond tokens, that still tend
      to be position latents. To get a latent-wise representative
      measure, we averaged the Fisher's Z-transformed $\rho$
      across the set of molecules in the validation set,
      transformed them back to Spearman coefficients $\bar{\rho}$,
      and labeled those with $|\bar{\rho}| \geq 0.6$ as
      position latents.
      
      \textbf{Anything-But-X-Latents.} Since these latents are
      expected to be activated on any token, we argue that such
      latents have dense activation patterns (activation density
      $\geq 0.2$) and distinct non-activated tokens. Activation
      density was measured by averaging the proportion of
      activated tokens per SMILES string per latent across the
      validation set. For distinct non-activated tokens, we
      used $p_{\text{ctx}}$ calculated over the validation set
      as previously described in \cref{main:anything_but_x},
      with a cutoff of $\geq 0.9$ for tokens with total
      occurrence $\ge 100$.

  \subsection{Chemically Relevant Latents}
    \textbf{Datasets.} The datasets for each task were fetched
    using the pytdc package. Most of the 21 tasks used in
    the study were split using scaffold split, while certain
    tasks---"Lipophilicity (AstraZeneca)", "Hydration Free
    Energy (FreeSolv)", "Tox21", and "ClinTox"---used random
    split. For each task, three different data splits were
    produced using different random seeds with 80/20
    training-test set split.

    \textbf{Probe Features.} As baselines, we used
    Extended-Connectivity Fingerprints (ECFPs; i.e., Morgan
    Fingerprints) with diameters of 0, 2, and 4, along with
    12 physicochemical descriptors: (1) Crippen's octanol-water
    partition coefficient and (2) molar refractivity, (3)
    molecular weight, (4) formal charge, (5) fraction of $sp^3$
    hybridized carbon, (6) heavy atom count, (7) total atom
    count (including implicit hydrogens), (8) ring count,
    (9) number of hydrogen bond acceptors and (10) donors,
    (11) number of rotatable bonds, and (12) topological polar
    surface area (TPSA); all generated using RDKit. The
    physicochemical descriptors were standardized task-wise
    over the training set and the mean and variance were
    reused to standardize the test set features. For
    MolFormer-10 hidden states and SAE activations across
    layers, features were obtained via mean- and max-pooling
    over the SMILES token sequence.

    \textbf{Linear Probes.} We trained logistic regression and
    ridge regression for classification and regression tasks,
    respectively. All linear probes were optimized using grid
    search over regularization strengths ($10^{-4}$, $10^{-2}$,
    $10^{0}$, $10^{2}$, $10^{4}$) with 5-fold cross-validation
    and evaluated on a separate test set for each data split.

\clearpage

\section{Additional Results}
  \subsection{Syntax Parsing Features}  
    \vfill
    \begin{figure*}[htbp]
      \centering
      \includegraphics[width=0.98\textwidth]{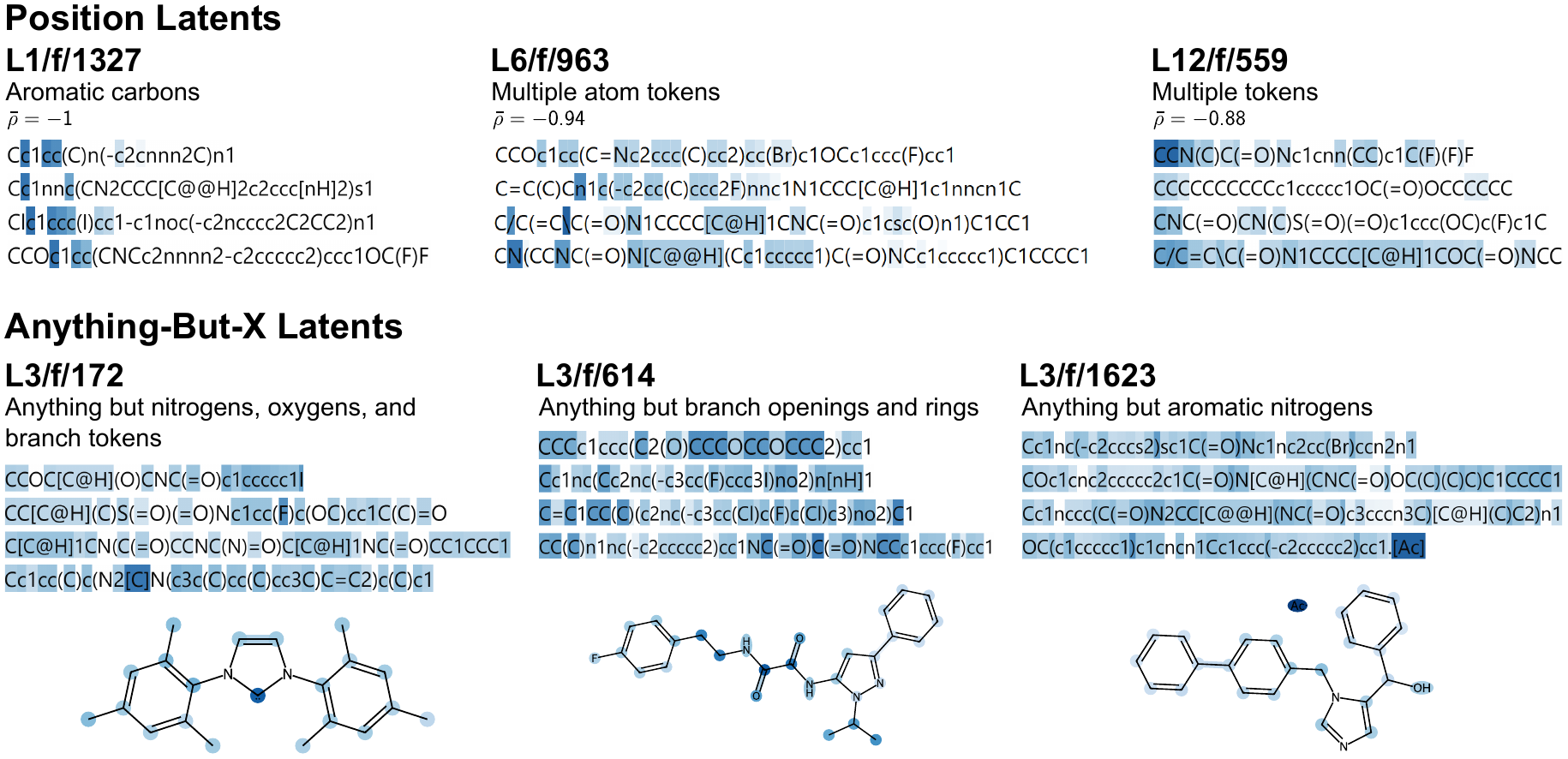}
      \caption{
        \textbf{More Examples of Syntax Parsing Features.}
        \textbf{Position Latents}: \textbf{L1/f/1327} shows
        decreasing activations on token \texttt{c} in the first
        few positions; \textbf{L3/f/1803} shows decreasing
        activations on multiple atom tokens; and \textbf{L12/f/559}
        shows decreasing activations across multiple tokens.
        \textbf{Anything-But-X Latents}: \textbf{L3/f/172}
        ($p_{\text{zero}}=0.97$) does not activate on nitrogens
        ($p_{\text{ctx}}=0.99$), oxygens ($p_{\text{ctx}}=0.99$),
        branch openings ($p_{\text{ctx}}=0.99$), and closings
        ($p_{\text{ctx}}=0.99$); \textbf{L3/f/614}
        ($p_{\text{zero}}=0.98$) does not activate on branch
        openings ($p_{\text{ctx}}=0.99$) and rings
        ($p_{\text{ctx}}=0.99$); and \textbf{L3/f/1623}
        ($p_{\text{zero}}=0.59$) does not activate on aromatic
        nitrogens ($p_{\text{ctx}}=0.99$).
      }
      \label{fig:apx_syntax_features}
    \end{figure*}
    \vfill
    \clearpage

  \subsection{SMILES Variants Perturb Representations}
    \vfill
    \begin{figure*}[htbp]
      \centering
      \includegraphics[width=\textwidth]{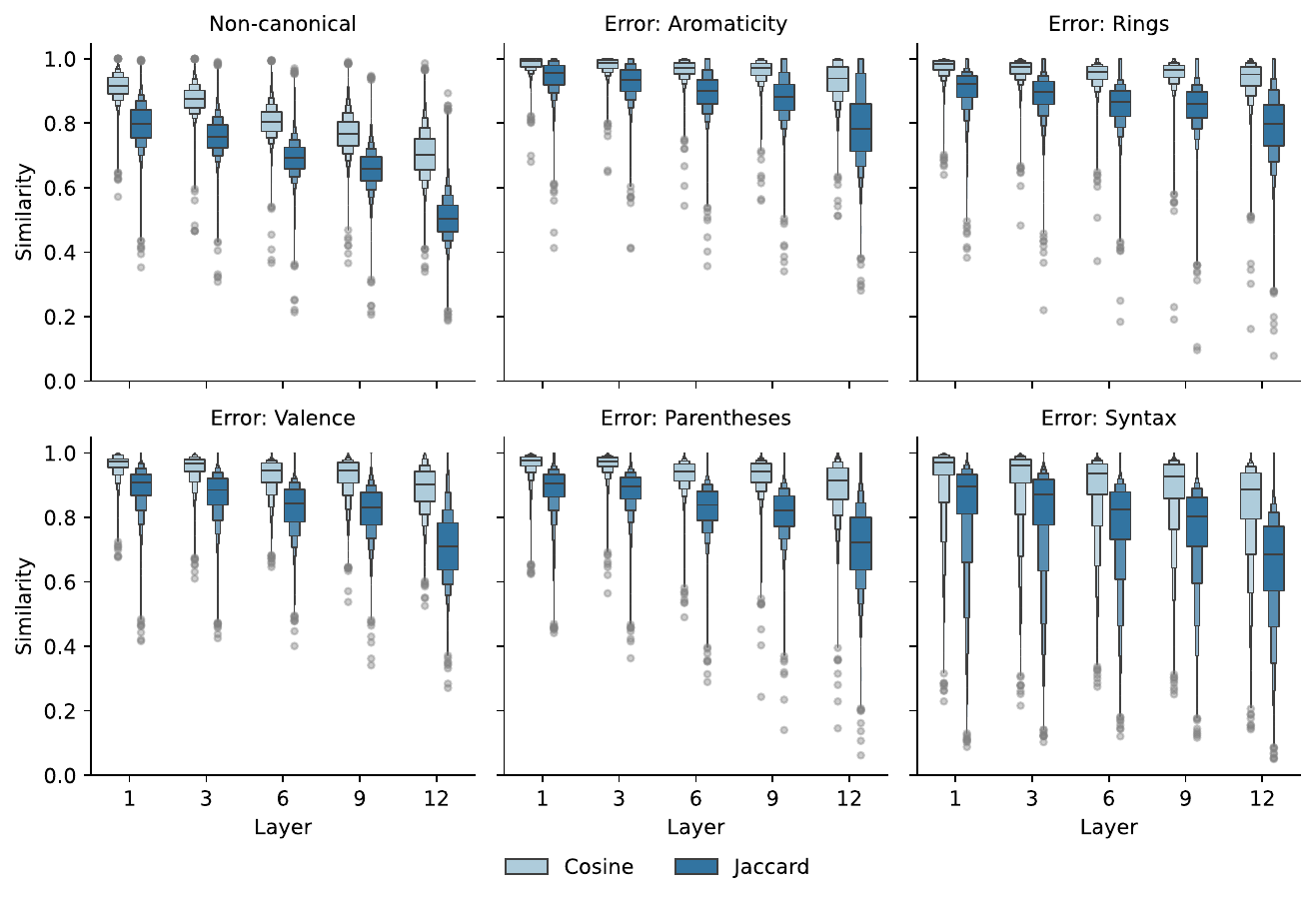}
      \caption{
        \textbf{Distribution of Similarity Measures for Non-canonical
        and Invalid SMILES.} Generally, we find that the median
        Jaccard and cosine similarities for invalid SMILES are
        higher compared to non-canonical SMILES, suggesting that
        latent representations are more robust to invalid SMILES
        than to non-canonical SMILES. This is likely because
        invalid SMILES retain the canonical token order, whereas
        non-canonical SMILES reorder the atomic traversal and
        thus shuffle the token sequence. When sorted by median
        Jaccard and cosine similarity from most to least similar
        to valid molecules, aromaticity errors are the most
        similar up to Layer 9, while ring errors being the most
        similar at the Layer 12. Valence and parentheses errors
        are typically the third and fourth most similar, with
        syntax errors being the most perturbed.
      }
      \label{fig:suppl_sim}
    \end{figure*}
    \vfill

    \begin{figure*}[htbp]
      \centering
      \includegraphics[width=\textwidth]{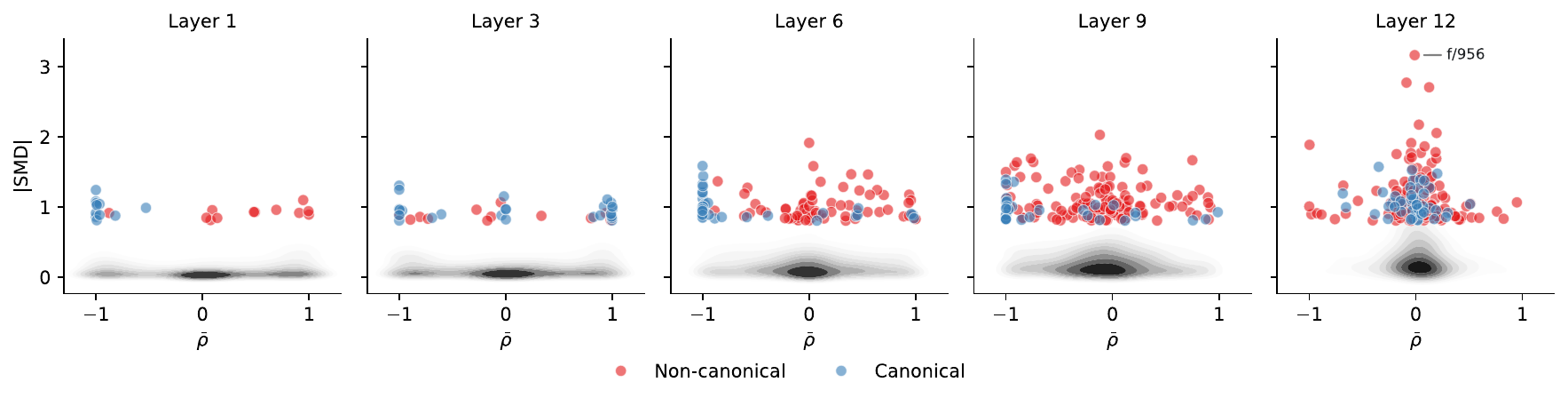}
      \caption{
        \textbf{Non-canonical SMILES: Significant Latents
        Increase Across Layers.} Grey contour areas show the
        distribution of latent $|\text{SMD}|$ and Spearman
        correlation to token position $\bar{\rho}$; colored
        points mark significant latents ($|\text{SMD}| \geq 0.8$)
        from max-pooled activations, with blue indicating
        higher $|\text{SMD}|$ for canonical and red for
        non-canonical SMILES. Canonical latents consistently
        cluster near $\bar{\rho} \approx -1.0$, indicating
        position latents, but predominate only in early layers.
        From Layer 6 to Layer 12, the number of significant
        non-canonical latents exceeds that of canonical ones,
        likely reflecting how disruption of early-layer position
        latents alters internal molecular representations in
        deeper layers.
      }
      \label{fig:suppl_nc}
    \end{figure*}

    \begin{figure*}[htbp]
        \centering
        \includegraphics[width=\textwidth]{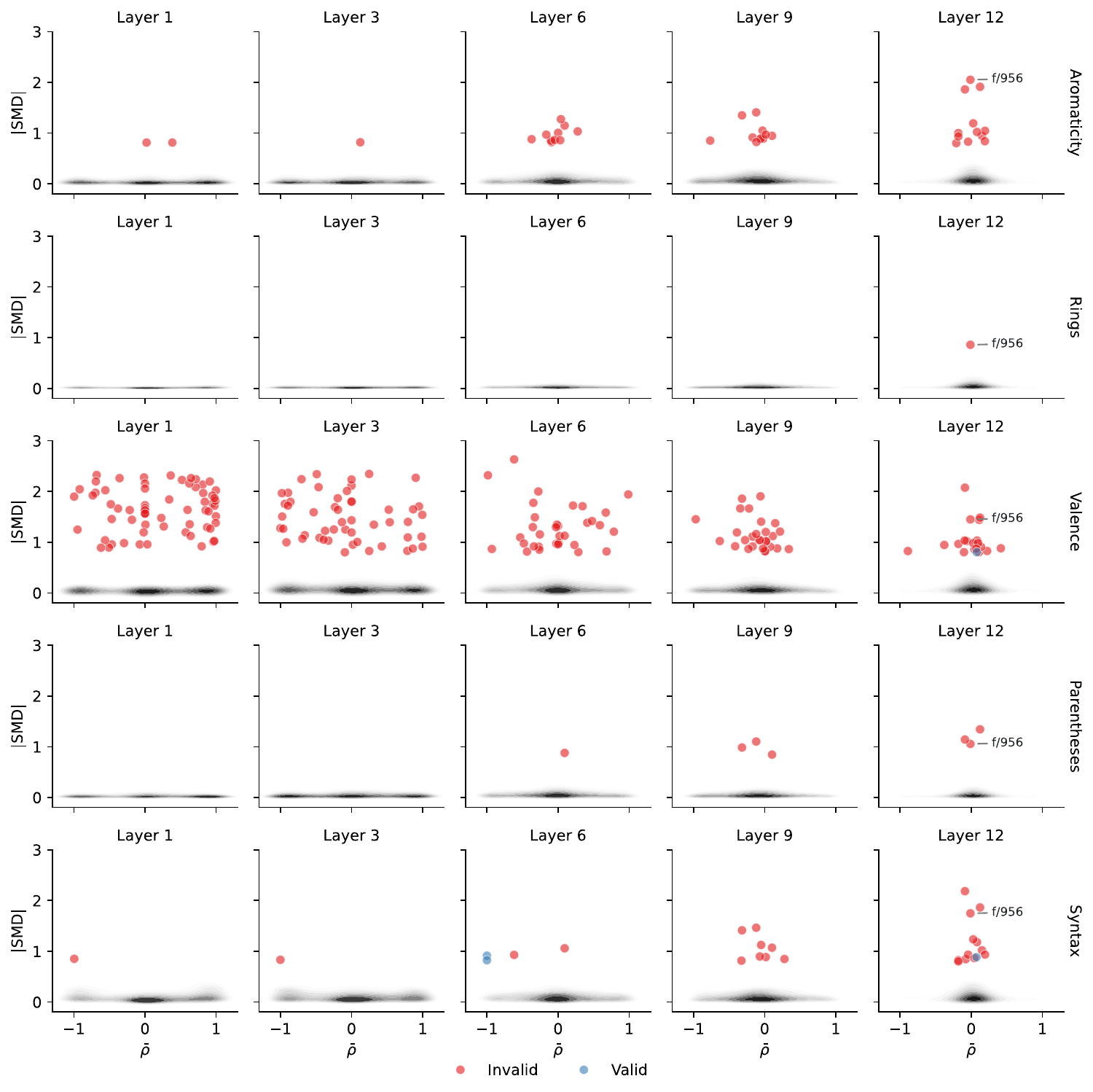}
        \caption{
          \textbf{Invalid SMILES: Valence Errors Have the Most
          Significant Latents in Early Layers.} Grey contour
          areas show the distribution of latent $|\text{SMD}|$
          and Spearman correlation to token position $\bar{\rho}$;
          colored points mark significant latents
          ($|\text{SMD}| \geq 0.8$) from max-pooled activations,
          with blue indicating higher $|\text{SMD}|$ for valid
          and red for invalid SMILES. Unlike augmented SMILES
          case, high-SMD latents across all error types are
          mainly driven by stronger activations in invalid SMILES.
          Interestingly, valence errors have the greatest number
          of high-SMD latents but are the only error type whose
          count decreases through layer 12, which may be attributed
          to how valence errors introduce or modify bond tokens,
          mostly affecting early-layer latents \cref{tab:errors}.
        }
      \label{fig:suppl_err}
    \end{figure*}
    \clearpage

  \subsection{Causal Evidence of Position Latent Propagation}
    We hypothesize that the low final-layer embeddings
    similarity observed for non-canonical SMILES arises
    from errors propagating through position latents in
    early layers. To test this, we ablate these latents
    by patching their activations with values from the
    canonical variant, which should restore embedding
    consistency between non-canonical and canonical
    SMILES, both representing the same molecule.
  
    \textbf{Dataset.} We use the same dataset as in
    the augmented SMILES analysis. However, since this
    analysis requires patching activations from canonical
    SMILES into their non-canonical counterparts, we
    filter for non-canonical SMILES with the same token
    length as their canonical ones. This results in
    a smaller set of 12,186 canonical SMILES, with a
    total of 21,545 pairs.

\begin{figure*}[htbp]
      \centering
      \includegraphics[width=\textwidth]{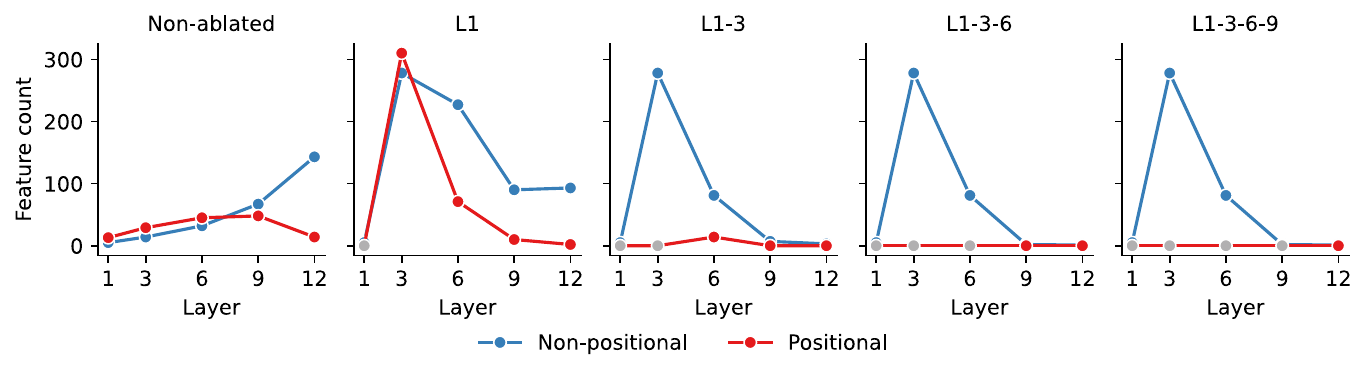}
      \caption{
        \textbf{Early-Layer Position Latents Affect Final-Layer
        Molecular Representations.} The y-axis represents the
        number of SAE latents with $|\text{SMD}| \geq 0.8$, and
        the grey markers indicate the ablated position latents,
        for which the count is expected to be zero. Activation
        patching reverses this trend, with layer 3 surging
        significantly for both positional and non-positional
        latents upon ablating layer 1. Ablating layers 1 and 3
        significantly reduces high-SMD non-positional latents
        at layer 9 and 12; adding layer 6 drops these further
        to 1 and 0, respectively. This shows that ablating
        layers 1 and 3 both reduces downstream high-SMD position
        latents and corrects later layers' molecular embeddings.
        Notably, the surge appears at layer 3 and only after
        ablating layer 1, suggesting layer 3 plays a key role
        in correcting MolFormer's internal representations.
      }
      \label{fig:suppl_ncablatepos}
    \end{figure*}

    \textbf{Activation Patching.} Activations are patched
    cumulatively across four configurations: Layer 1;
    Layer 1 and 3; Layer 1, 3, and 6; and Layer 1, 3, 6,
    and 9. Unchanged activations are referred to as
    non-ablated, as shown in \cref{fig:suppl_ncablatepos}.
    In analyzing SAE latent dynamics across layers, we
    use SMD (see \cref{main:alt_repr}) for each latent,
    computed from its max-pooled across tokens, molecule-wise.

  \begin{landscape}
  \subsection{Downstream Tasks}
    \vfill
    \begin{tabularx}{0.99\linewidth}{l*{4}{>{\centering\arraybackslash\hsize=0.85\hsize}X}*{4}{>{\centering\arraybackslash\hsize=1.15\hsize}X}}
    \caption{
        \textbf{Performance on ADMET Classification Tasks.} Values are reported as the mean
        $\pm$ standard deviation of ROC-AUC ($\uparrow$) across three runs with different
        random seeds for train-test splits. Tox21 task results are averaged across 12
        different targets. Different pooling methods are used to generate MolFormer and SAE
        molecular features.
    }
    \label{tab:admet_cls} \\
    \toprule
        \multirow{2}{*}{Task} & \multicolumn{4}{c}{Baseline} & \multicolumn{2}{c}{MolFormer} & \multicolumn{2}{c}{SAE} \\
        & ECFP0 & ECFP2 & ECFP4 & PcDs\footnotemark[1] & Mean & Max & Mean & Max \\
    \midrule
    \endfirsthead

    \toprule
        \multirow{2}{*}{Task} & \multicolumn{4}{c}{Baseline} & \multicolumn{2}{c}{MolFormer} & \multicolumn{2}{c}{SAE} \\
        & ECFP0 & ECFP2 & ECFP4 & PcDs\footnotemark[1] & Mean & Max & Mean & Max \\
    \midrule
    \endhead

    \midrule
    \endfoot

    \bottomrule
    \endlastfoot
        \multicolumn{1}{l}{\textbf{Absorption (A)}} \\
            Bioavailability (Ma) & 0.63 $\pm$ 0.07 & 0.66 $\pm$ 0.09 & 0.68 $\pm$ 0.08 & 0.68 $\pm$ 0.08 & 0.71 $\pm$ 0.05 (12) & \textbf{0.74} $\bm{\pm}$ \textbf{0.07} \textbf{(9)} & \underline{0.72 $\pm$ 0.04 (12)} & 0.71 $\pm$ 0.08 (12) \\
            HIA (Hou) & 0.95 $\pm$ 0.03 & 0.97 $\pm$ 0.01 & 0.96 $\pm$ 0.02 & \textbf{1.0} $\bm{\pm}$ \textbf{0.0} & 0.97 $\pm$ 0.01 (12) & 0.97 $\pm$ 0.02 (3) & 0.98 $\pm$ 0.01 (12) & \underline{0.98 $\pm$ 0.02 (9)} \\
            PAMPA (NCATS) & 0.66 $\pm$ 0.04 & 0.71 $\pm$ 0.03 & 0.72 $\pm$ 0.01 & 0.75 $\pm$ 0.01 & 0.75 $\pm$ 0.02 (12) & \underline{0.75 $\pm$ 0.02 (12)} & \textbf{0.76} $\bm{\pm}$ \textbf{0.02} \textbf{(12)} & 0.75 $\pm$ 0.03 (12) \\
            Pgp (Broccatelli) & 0.77 $\pm$ 0.06 & 0.91 $\pm$ 0.03 & 0.91 $\pm$ 0.02 & 0.91 $\pm$ 0.0 & 0.92 $\pm$ 0.01 (6) & \underline{0.92 $\pm$ 0.02 (12)} & 0.92 $\pm$ 0.0 (1) & \textbf{0.93} $\bm{\pm}$ \textbf{0.01} \textbf{(12)} \\ [18px]
        \multicolumn{1}{l}{\textbf{Distribution (D)}} \\
            BBB (Martins) & 0.81 $\pm$ 0.04 & 0.88 $\pm$ 0.03 & 0.86 $\pm$ 0.05 & 0.81 $\pm$ 0.01 & 0.87 $\pm$ 0.02 (6) & 0.86 $\pm$ 0.01 (12) & \underline{0.88 $\pm$ 0.02 (6)} & \textbf{0.9} $\bm{\pm}$ \textbf{0.03} \textbf{(6)} \\ [18px]
        \multicolumn{1}{l}{\textbf{Metabolism (M)}} \\
            CYP2C9 (Veith) & 0.73 $\pm$ 0.03 & 0.86 $\pm$ 0.01 & 0.86 $\pm$ 0.01 & 0.8 $\pm$ 0.01 & 0.86 $\pm$ 0.01 (12) & 0.87 $\pm$ 0.01 (12) & \underline{0.87 $\pm$ 0.01 (12)} & \textbf{0.88} $\bm{\pm}$ \textbf{0.01} \textbf{(12)} \\
            CYP2D6 (Veith) & 0.75 $\pm$ 0.02 & 0.85 $\pm$ 0.0 & 0.84 $\pm$ 0.01 & 0.77 $\pm$ 0.01 & 0.85 $\pm$ 0.0 (6) & 0.84 $\pm$ 0.01 (12) & \textbf{0.86} $\bm{\pm}$ \textbf{0.0} \textbf{(6)} & \underline{0.85 $\pm$ 0.01 (12)} \\ [18px]
        \multicolumn{1}{l}{\textbf{Toxicity (T)}} \\
            AMES & 0.7 $\pm$ 0.02 & 0.81 $\pm$ 0.01 & 0.79 $\pm$ 0.01 & 0.73 $\pm$ 0.01 & 0.8 $\pm$ 0.01 (12) & 0.81 $\pm$ 0.0 (12) & \underline{0.82 $\pm$ 0.01 (12)} & \textbf{0.82} $\bm{\pm}$ \textbf{0.02} \textbf{(9)} \\
            Carcinogens (Lagunin) & 0.73 $\pm$ 0.17 & 0.72 $\pm$ 0.22 & 0.72 $\pm$ 0.21 & 0.73 $\pm$ 0.16 & 0.74 $\pm$ 0.07 (1) & \underline{0.75 $\pm$ 0.11 (3)} & 0.74 $\pm$ 0.19 (1) & \textbf{0.75} $\bm{\pm}$ \textbf{0.13} \textbf{(1)} \\
            ClinTox & 0.9 $\pm$ 0.04 & 0.87 $\pm$ 0.05 & 0.82 $\pm$ 0.07 & 0.77 $\pm$ 0.05 & 0.9 $\pm$ 0.03 (9) & \underline{0.91 $\pm$ 0.01 (12)} & 0.9 $\pm$ 0.03 (6) & \textbf{0.91} $\bm{\pm}$ \textbf{0.02} \textbf{(6)} \\
            Tox21\footnotemark[2] & 0.74 $\pm$ 0.04 & 0.81 $\pm$ 0.05 & 0.79 $\pm$ 0.05 & 0.76 $\pm$ 0.05 & 0.81 $\pm$ 0.05 (9) & 0.81 $\pm$ 0.04 (9) & \textbf{0.82} $\bm{\pm}$ \textbf{0.05} \textbf{(9)} & \underline{0.81 $\pm$ 0.05 (12)} \\
\end{tabularx}

\footnotetext[1]{Physicochemical descriptors.}
\footnotetext[2]{Reported scores are also averaged across multiple binary targets.}

    \vfill
    \clearpage

    \vspace*{\fill}
    \begin{tabularx}{0.99\linewidth}{>{\RaggedRight\arraybackslash\hyphenpenalty=10000}p{3.25cm}*{4}{>{\centering\arraybackslash\hsize=0.9\hsize}X}*{4}{>{\centering\arraybackslash\hsize=1.1\hsize}X}}
    \caption{
        \textbf{Performance on ADMET Regression Tasks.} Values are reported as the mean
        $\pm$ standard deviation of RMSE ($\downarrow$) across three runs with different
        random seeds for train-test splits. Different pooling methods are used to
        generate MolFormer and SAE molecular features.
    }
    \label{tab:admet_reg} \\
    \toprule
        \multirow{2}{*}{Task} & \multicolumn{4}{c}{Baseline} & \multicolumn{2}{c}{MolFormer} & \multicolumn{2}{c}{SAE} \\
        & ECFP0 & ECFP2 & ECFP4 & PcDs\footnotemark[1] & Mean & Max & Mean & Max \\
    \midrule
    \endfirsthead
    
    \toprule
        \multirow{2}{*}{Task} & \multicolumn{4}{c}{Baseline} & \multicolumn{2}{c}{MolFormer} & \multicolumn{2}{c}{SAE} \\
        & ECFP0 & ECFP2 & ECFP4 & PcDs\footnotemark[1] & Mean & Max & Mean & Max \\
    \midrule
    \endhead
    
    \midrule
    \endfoot
    
    \bottomrule
    \endlastfoot
        \multicolumn{1}{l}{\textbf{Absorption (A)}} \\
            Caco2 (Wang) & 0.66 $\pm$ 0.06 & 0.59 $\pm$ 0.03 & 0.56 $\pm$ 0.02 & 0.54 $\pm$ 0.01 & \textbf{0.49} $\bm{\pm}$ \textbf{0.02} \textbf{(12)} & 0.52 $\pm$ 0.06 (9) & 0.51 $\pm$ 0.03 (12) & \ul{0.5 $\pm$ 0.0 (12)} \\
            Hydration Free Energy (FreeSolv) & 2.26 $\pm$ 0.36 & 1.39 $\pm$ 0.34 & 1.6 $\pm$ 0.37 & 1.5 $\pm$ 0.05 & \ul{1.38 $\pm$ 0.08 (3)} & 1.51 $\pm$ 0.12 (1) & 1.53 $\pm$ 0.23 (6) & \textbf{1.2} $\bm{\pm}$ \textbf{0.19} \textbf{(1)} \\
            Lipophilicity (AstraZeneca) & 1.11 $\pm$ 0.07 & 0.81 $\pm$ 0.03 & 0.85 $\pm$ 0.05 & 1.03 $\pm$ 0.06 & 0.78 $\pm$ 0.04 (12) & 0.81 $\pm$ 0.03 (12) & \textbf{0.75} $\bm{\pm}$ \textbf{0.04} \textbf{(12)} & \ul{0.75 $\pm$ 0.04 (12)} \\
            Solubility (AqSolDB) & 2.39 $\pm$ 0.07 & 2.02 $\pm$ 0.21 & 1.87 $\pm$ 0.13 & 2.21 $\pm$ 0.27 & 1.88 $\pm$ 0.59 (3) & \ul{1.77 $\pm$ 0.41 (1)} & 1.83 $\pm$ 0.57 (6) & \textbf{1.65} $\bm{\pm}$ \textbf{0.35} \textbf{(1)} \\ [18 px]
        \multicolumn{1}{l}{\textbf{Distribution (D)}} \\
            PPBR (AZ) & 14.95 $\pm$ 1.61 & 13.4 $\pm$ 1.46 & 13.25 $\pm$ 1.38 & 13.72 $\pm$ 1.28 & 12.79 $\pm$ 0.76 (12) & 13.3 $\pm$ 1.15 (6) & \ul{12.75 $\pm$ 1.05 (9)} & \textbf{12.66} $\bm{\pm}$ \textbf{0.99} \textbf{(6)} \\
            VDss (Lombardo) & 20.7 $\pm$ 23.56 & 20.55 $\pm$ 23.41 & 20.54 $\pm$ 23.44 & 20.44 $\pm$ 23.66 & 20.44 $\pm$ 23.75 (12) & \textbf{20.26} $\bm{\pm}$ \textbf{23.49} \textbf{(6)} & 20.34 $\pm$ 23.54 (6) & \ul{20.32 $\pm$ 23.64 (1)} \\ [18px]
        \multicolumn{1}{l}{\textbf{Excretion (E)}} \\
            Clearance Hepatocyte (AZ) & 49.11 $\pm$ 1.87 & 46.78 $\pm$ 1.3 & 46.57 $\pm$ 0.54 & 49.8 $\pm$ 1.76 & \ul{46.47 $\pm$ 0.5 (12)} & 47.11 $\pm$ 1.2 (1) & 46.51 $\pm$ 1.81 (9) & \textbf{45.58} $\bm{\pm}$ \textbf{1.25} \textbf{(12)} \\
            Clearance Microsome (AZ) & 43.13 $\pm$ 1.63 & 39.08 $\pm$ 1.88 & 39.6 $\pm$ 1.86 & 44.14 $\pm$ 1.42 & 39.02 $\pm$ 2.12 (12) & 39.89 $\pm$ 1.05 (9) & \ul{38.49 $\pm$ 1.63 (12)} & \textbf{37.43} $\bm{\pm}$ \textbf{2.06} \textbf{(12)} \\
            Half Life (Obach) & 23.74 $\pm$ 5.2 & 29.38 $\pm$ 5.87 & 30.1 $\pm$ 5.08 & 38.57 $\pm$ 30.37 & \ul{19.81 $\pm$ 2.4 (12)} & \textbf{19.66} $\bm{\pm}$ \textbf{2.91} \textbf{(6)} & 26.0 $\pm$ 8.99 (9) & 25.64 $\pm$ 8.21 (3) \\ [18px]
        \multicolumn{1}{l}{\textbf{Toxicity (T)}} \\
            LD50 (Zhu) & 0.85 $\pm$ 0.03 & 0.81 $\pm$ 0.06 & \ul{0.79 $\pm$ 0.02} & 0.91 $\pm$ 0.02 & 0.81 $\pm$ 0.02 (6) & 0.83 $\pm$ 0.03 (1) & 0.8 $\pm$ 0.01 (6) & \textbf{0.78} $\bm{\pm}$ \textbf{0.02} \textbf{(6)} \\
\end{tabularx}

\footnotetext[1]{Physicochemical descriptors.}

    \vspace*{\fill}
    \clearpage
  \end{landscape}

    \begin{minipage}[c][\textheight]{\textwidth}
      \centering
      \includegraphics[width=\textwidth]{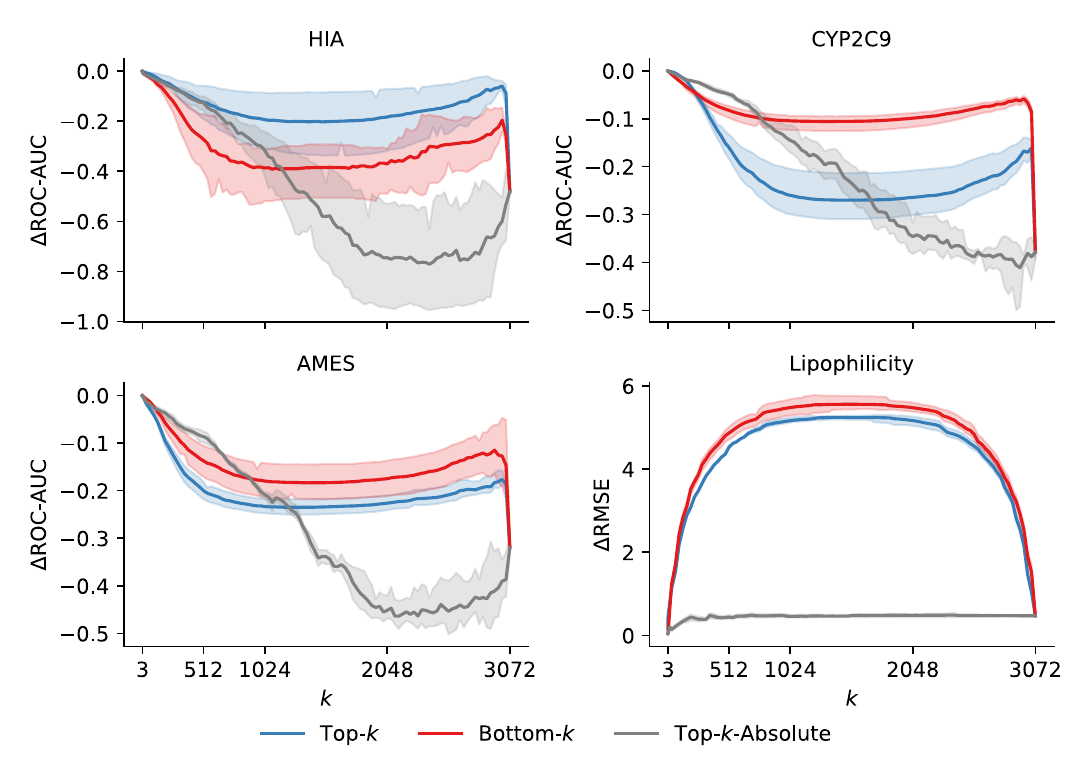}
      \captionof{figure}{
        \textbf{Pharmacological Encoding Is Shared Across
        Multiple SAE Latents.} Shown here is the change
        in linear probe performance upon ablating latents
        ranked by $z(\beta_{f})$ -- top-$k$, bottom-$k$,
        and top-$k$ by absolute value -- on the test set
        of each data split. Ablating the top-3, bottom-3,
        or even top-3-absolute latents does not significantly
        degrade performance, whereas ablating 512 or more
        latents does, indicating that task encoding is
        distributed across many SAE latents. Across both
        top-$k$ and bottom-$k$ ablations, classification
        and regression tasks show similar performance
        degradation trends, except that classification
        tasks exhibit a slight recovery at very large $k$.
        Meanwhile, the regression task produces symmetric
        $\Delta$RMSE curves, since $\Delta$RMSE is evaluated
        in absolute value; the top-$k$-absolute curve still
        suggests similar behavior to classification probes.
      }
      \label{fig:suppl_taskablate}
    \end{minipage}

    \begin{minipage}[c][\textheight]{\textwidth}
      \centering
      \includegraphics[width=0.98\textwidth]{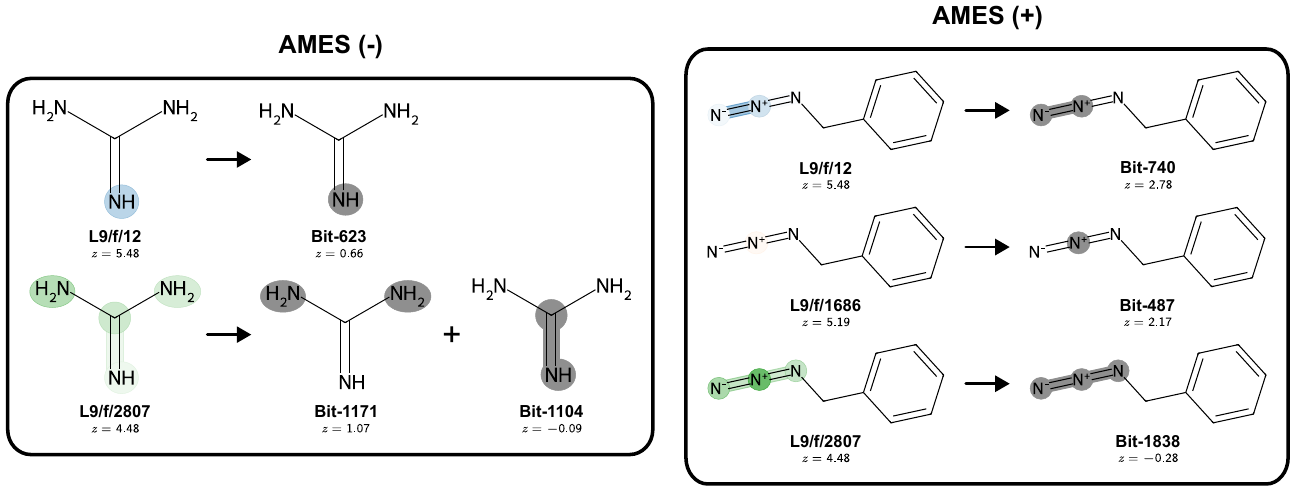}
      \captionof{figure}{
        \textbf{SAE Latents Align with Morgan Fingerprint
        Features.} Two molecules from the AMES dataset show
        that concepts encoded in the SAE latents can
        correspond to ECFP2 bits. A single SAE latent may
        be correspond to a fraction of, exactly one, or
        many ECFP2 bits. Moreover, SAE latents offer not
        only binary representations but also activation
        magnitudes, as shown by L9/f/2807, which activates
        strongly at the \texttt{[N+]} atom in the AMES (+)
        molecule. This richer representation may also explain
        the slightly higher evaluation performance of SAE
        latents, particularly when using max-pooling features,
        over ECFP on most downstream tasks.
      }
      \label{fig:suppl_ecfp2}
    \end{minipage}


\end{document}